%% file: main.tex
\documentclass[11pt,a4paper]{article}
\usepackage[hyperref]{acl2020}
\usepackage{times}
\usepackage{latexsym}
\usepackage{pgfplots}
\usepackage{booktabs}
\usepackage{diagbox}
\pgfplotsset{width=.9\columnwidth}
\usepackage{url}

\aclfinalcopy 

\usepackage{amsmath}
\usepackage[colorinlistoftodos]{todonotes}
\usepackage{multirow}
\usepackage{bbding}
\usepackage{subfigure}
\usepackage{CJKutf8}

\newcommand\ttt[1]{\texttt{#1}}

\date{}

\title{AltCLIP: Altering the Language Encoder in CLIP for Extended Language Capabilities}

\author{
Zhongzhi Chen \textsuperscript{*1,2\dag},\
Guang Liu  \thanks{~ Equal Contribution} \textsuperscript{ 1},\
Bo-Wen Zhang\textsuperscript{1}, \
Fulong Ye\textsuperscript{1,3\dag} \thanks{~ Work done during internship with Beijing Academy of Artificial Intelligence} \\
\bf{Qinghong Yang\textsuperscript{2}, \
Ledell Wu\textsuperscript{1}}  \\
\textsuperscript{1} Beijing Academy of Artificial Intelligence, \
\textsuperscript{2} Beihang University, \\
\textsuperscript{3} Beijing University of Posts and Telecommunications.\\
\texttt{\{liuguang, wuyu\}@baai.ac.cn} 
}

\pgfplotsset{compat=1.17}
\begin{document}
\maketitle
\begin{abstract}
In this work, we present a conceptually simple and effective method to train a strong bilingual/multilingual multimodal representation model. Starting from the pre-trained multimodal representation model CLIP released by OpenAI, we altered its text encoder with a pre-trained multilingual text encoder XLM-R, and aligned both languages and image representations by a two-stage training schema consisting of teacher learning and contrastive learning. We validate our method through evaluations of a wide range of tasks. We set new state-of-the-art performances on a bunch of tasks including ImageNet-CN, Flicker30k-CN, COCO-CN and XTD. Further, we obtain very close performances with CLIP on almost all tasks, suggesting that one can simply alter the text encoder in CLIP for extended capabilities such as multilingual understanding. Our models and code are available at \url{https://github.com/FlagAI-Open/FlagAI}.

\end{abstract}

\input{1_introduction.tex}

\input{2_related_work.tex}
\input{3_method.tex}
\input{4_model_training.tex}
\input{5_experiments.tex}
\input{6_conclusion.tex}

\input{7_acknowledgement.tex}

\bibliography{reference}
\bibliographystyle{acl_natbib}
\appendix
\input{A_appendix.tex}

\end{document}

%% file: 1_introduction.tex
\section{Introduction}
Learning a good representation in a joint space for vision and language has been a long pursuit in the research of Artificial Intelligence (AI). Recently, the milestone work of CLIP \cite{radford2021learning} from OpenAI demonstrated impressive zero-shot performances across a number of tasks such as image classification on ImageNet~\cite{deng2009imagenet}, Image-to-Text and Text-to-Image retrieval on Flicker-30k ~\cite{young2014image} and MSCOCO\cite{lin2014microsoft, chen2015microsoft}. There has been the pursuit of building contrastive language-image models in other languages such as Italian ~\cite{bianchi2021contrastive}, Korean ~\cite{ko2022large}, Chinese ~\cite{changpinyo2021conceptual, fei2021wenlan, fengshenbang, gu2022wukong, xie2022zero, yang2022chinese} or in a cross-lingual and multilingual setting ~\cite{aggarwal2020towards, carlsson2022cross}.

Training a good language-image representation model often requires a huge amount of text-image pairs and vast computational resources. For instance, CLIP used 400M text-image pairs, and Taiyi ~\cite{fengshenbang}, a recently proposed Chinese model, used 123M text-image pairs. To alleviate this problem, there are multiple works that try to utilize existing pretrained models and only pretrain part of the network ~\cite{portaz2019image, aggarwal2020towards, gu2022wukong, zhai2022lit}. More recently, \newcite{carlsson2022cross} proposed to use Teacher Learning (a.k.a. Knowledge Distillation) on the text encoder of the CLIP model to learn a multilingual text-image representation model. This method only uses machine-translated data from English to a target language, without text-to-image pairs. 

However, existing works in the cross-lingual or multilingual setting mainly focus on the model's retrieval performance and ignores their generalization ability. The data set to evaluate retrieval performance is often small, e.g., $1,000$ images in test sets for Flickr-30k. The retrieval performance fluctuates acutely with the change in training data distribution. Although current methods achieve good performance in retrieval, these methods often do not perform well on the ImageNet classification tasks. The ability to accurately predict images over $1,000$ classes often indicates better generalization ability of the model. 

\begin{figure*}[t]
\centering
\includegraphics[width=1\textwidth]{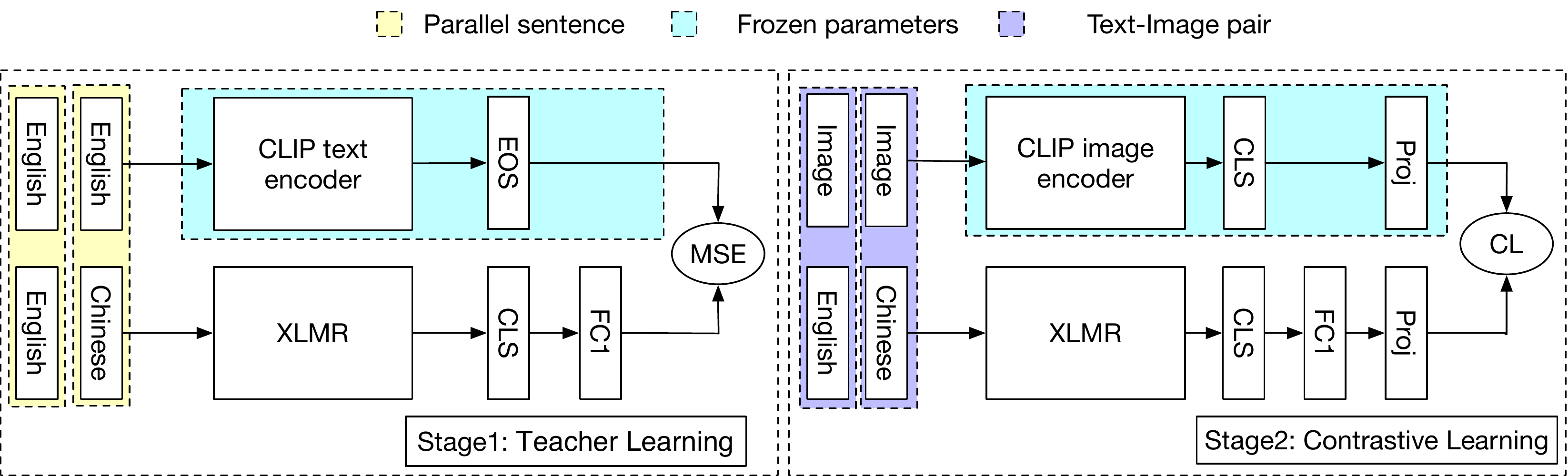}
\caption{\label{fig:framework}The framework of our method.}
\end{figure*}

To address the aforementioned problems, we propose a bilingual model named Alter ego CLIP (\textbf{AltCLIP}) which achieved strong performances on  ImageNet and multimodal retrieval tasks in both English and Chinese. Our AltCLIP learns a strong bilingual language-image representation under a two-stage framework (see Figure \ref{fig:framework} for an overview). In the first stage, we use Teacher Learning to distill the knowledge learned from CLIP. In the second stage, we train the model via Contrastive Learning ~\cite{hadsell2006dimensionality} on a relatively small amount of Chinese and English text-image pairs. We show the effectiveness of our method by experimenting with a wide range of benchmarks in English and Chinese. Further, We set new state-of-the-art results on multiple image classification and retrieval tasks in Chinese. We further extended this method to train a multilingual multimodal model where we call it AltCLIP$_{M9}$. The AltCLIP$_{M9}$ model achieves state-of-the-art zero-shot results on the multilingual text-image retrieval dataset XTD \cite{aggarwal2020zeroshot}.


%% file: 2_related_work.tex
\section{Related Work}
CLIP \cite{radford2021learning} provides a strong English text-image representation. To expand the language of CLIP model,
there are prior studies on learning a bilingual text-image representation ~\cite{ko2022large, bianchi2021contrastive}, and multilingual text-image representation ~\cite{aggarwal2020towards, carlsson2022cross}. In the domain of Chinese text-image pretraining models, prior work such as Taiyi ~\cite{fengshenbang}, CN-CLIP ~\cite{yang2022chinese}, Wukong ~\cite{gu2022wukong}, R2D2 ~\cite{xie2022zero} and BriVL ~\cite{huo2021wenlan, fei2021wenlan}. These methods often need large-scale Chinese text-image pairs and suffer from a significant performance decline in English tasks.

\newcite{carlsson2022cross} proposed a way to utilize Teacher Learning (a.k.a. Knowledge Distillation) \cite{hinton2015distilling} to train a new textual encoder from the original CLIP model with only machine-translated parallel data, without using text-image pairs. Although this method achieves promising cross-lingual retrieval performances with only text data, its zero-shot classification performance in English drops significantly. We follow their work to learn a bilingual model from CLIP with a new text encoder, with the following changes: firstly, we use knowledge distillation on English text pairs in addition to machine-translated text pairs; secondly, we add human-curated translation data for better quality; lastly, we fine-tune the model with text-image pairs to further boost its performance.


XLM-R \cite{conneau2020unsupervised} is a multilingual language model that achieves strong performances on a wide range of cross-lingual tasks. In our work, we use the XLM-R model as the underlying text encoder and align it with the image encoder trained in CLIP, to achieve competitive performances on cross-lingual and cross-modality tasks. 

%% file: 3_method.tex
\section{Methodology} 
We propose a two-stage method to learn a good bilingual and multilingual language-image representation model. In the first stage, we follow the work of \newcite{carlsson2022cross} to use Teacher Learning to learn a multilingual text encoder from the CLIP text encoder. In this step, no image is needed in training and only language parallel data is used. In the second stage, we use text-image pairs to further fine-tune the model from contrastive learning.  Our overall training procedure is summarized in Figure ~\ref{fig:framework}.


\begin{table*}[htb]
\small
\centering
\begin{tabular}{clccccc} 
\hline
\multirow{2}{*}{Language} & \multicolumn{1}{c}{\multirow{2}{*}{Method}} & \multicolumn{5}{c}{Dataset}                                         \\ 
\cline{3-7}
                           & \multicolumn{1}{c}{}                         & ImageNet & ImageNet Sketch~ & ImageNet-A & ImageNet-R & ImageNetV2~  \\ 
\hline
\multirow{5}{*}{\rotatebox{90}{English}}   & CLIP                                         & \textbf{75.5}     & \textbf{59.6}             & \textbf{70.6}       & \textbf{87.9}       & \textbf{69.9}         \\
                           & M-CLIP                                       & 52.3     & 44.2             & 59.1       & 81.7       & 47.3         \\
                           & CN-CLIP                                      & 32.5     & 30.7             & 29.8       & 57.0         & 28.6         \\
                           & AltCLIP$_{T}$                                     & 74.7     & 59.2             & 70.4       & 87.9       & 68.8         \\
                           & AltCLIP                                      & 74.5     & 58.7             & 69.5       & 87.2       & 68.2         \\ 
\hline
\multirow{5}{*}{\rotatebox{90}{Chinese}}   & CLIP                                         & 1.9      & 1.7              & 4.7          & 4.4        & 1.8          \\
                           & M-CLIP                                       & 43.0       & 36.3             & 51.3       & 68.3       & 39.5         \\
                           & CN-CLIP                                      & 53.6     &47.5           & 42.8       & 78.1       & 47.8         \\
                           & AltCLIP$_{T}$                                     & 58.2     & 46.9             & \textbf{62.7}       & 82.1         & 53.3         \\
                           & AltCLIP                                      & \textbf{59.6}     &\textbf{48.4}             & 61.5       & \textbf{82.5}       &\textbf {54.0}         \\
\hline
\end{tabular}
\caption{Comparison Results of our proposed model and baseline models on image classification benchmarks, i.e. ImageNet and its variants. AltCLIP$_{T}$ and AltCLIP denotes our model after Teacher Learning Stage and after Contrastive Learning Stage, respectively. All image encoders used in these models are Vit-L for fair comparison. The metric reported is zero-shot classification accuracy.}
\label{tab:imagenet}
\end{table*}

\subsection{Teacher Learning Stage}
In this stage, we perform Teacher Learning ~\cite{hinton2015distilling} on text encoders. We use the text encoder from CLIP ~\cite{radford2021learning} as the teacher text encoder, and the XLM-R ~\cite{conneau2020unsupervised} model pretrained on multilingual data as the student encoder. A fully-connected layer is added to transform the output of the XLM-R model into the same output dimension as the teacher encoder. We use parallel text data in both English and Chinese to distill the knowledge of text-image alignment. 

Given parallel text input $(sent_1, sent_2)$, the teacher text encoder generates the learning target from input $sent_1$, which is the embedding of the \ttt{[TOS]} token, denoted by $x^{t}_{tos}$. The student text encoder generates embedding $x^{s}_{cls}$ from input $sent_2$. We minimize Mean Squared Error (MSE) between $x^{t}_{tos}$ and $x^{s}_{cls}$. 
After such training, the student text encoder can keep most of its multilingual capability and obtain text-image alignment capability in both languages. Note that the teacher encoder is only used at training time. At inference time, only the student encoder is used as the text encoder.

To show that our method is extensible at including more languages, we build a multilingual version that supports nine different languages: English(En), Chinese(Zh), Spanish(Es), French(Fr), Russian(Ru), Arabic(Ar), Japanese(Ja), Korean(Ko), and Italian(It). For the multilingual version, we align more languages with English, with the same concept and architecture as in the bilingual version.


\subsection{Contrastive Learning Stage}
This stage of training aims to further improve text-image alignment by contrastive learning on multilingual text-image pairs. As illustrated in Figure \ref{fig:framework}, here we use the image encoder from CLIP which is based on Vision Transformer (ViT) ~\cite{dosovitskiy2020image} as our image encoder, and use the student text encoder learned from the Teacher Learning Stage as our text encoder. 

We use Contrastive Loss ~\cite{hadsell2006dimensionality} between the output projection of the image encoder and text encoder, as done similarly in previous work ~\cite{radford2021learning}. We follow LiT~\cite{zhai2022lit} to freeze the image encoder at training time and only update the parameters in the text encoder. We observe that this stage of training further improves the  model's performance on various evaluation benchmarks, as presented in Section \ref{sec5}.

%% file: 4_model_training.tex
\section{Model Training}
\subsection{Training Datasets} \label{training data}
In this section, we describe the training datasets used in our two-stage model training. 

\paragraph{Teacher Learning Stage}
In this stage, we use parallel text corpus to align the original CLIP text encoder and our new text encoder initialized from XLM-R. Our parallel text corpus includes the following:
\begin{enumerate}

\item Machine translated data of text in Conceptual Captions (CC3M) ~\cite{sharma2018conceptual} and a 28M randomly sampled subset from Laion-400M ~\cite{schuhmann2021laion}. For the multilingual model, we use the same setting but fewer 10M subsets from Laion-400M for each language.
\item Human translated data of text from TSL2019 ~\cite{xu2019nlp}, a total number of 5M English to Chinese translations. For the multilingual model, we extracted the same amount for each language. Parallel data in different languages is extracted from the public database OPUS\cite{tiedemann2012parallel}\footnote{https://opus.nlpl.eu}. This type of parallel data provides more accurate translation.

\end{enumerate}



\paragraph{Contrastive Learning Stage}
We use high-quality text-image pair data in this stage. Note that multilingual data are only used in our multilingual version.
We use 2M bilingual text-image pairs for AltCLIP and 100M multilingual text-image pairs for the multilingual version, from the following:
\begin{enumerate}
\item Chinese text-image dataset from Wudao MM ~\cite{yuan2021wudaocorpora}, filtered by the Laion Aesthetic v2\footnote{https://github.com/christophschuhmann/improved-aesthetic-predictor} model with a threshold score over 5.5.
\item English text-image dataset from LAION 5B ~\cite{schuhmann2022laion}, a randomly sampled subset from the data filtered by the Laion Aesthstic v2 model with a threshold score over 6.

\item Multilingual text-image dataset from LAION Multilingual 2B ~\cite{schuhmann2022laion}, a randomly sampled subset from the data.
\end{enumerate}

\begin{table*}[hbt]
\centering
\small
\begin{tabular}{ccccccccc}\hline
\multirow{2}{*}{Dataset}   & \multirow{2}{*}{Method} & \multicolumn{3}{l}{Text-to-Image Retrival} & \multicolumn{3}{l}{Image-to-Text Retrival} & \multirow{2}{*}{MR} \\\cline{3-8}
                           &                         & R@1          & R@5          & R@10         & R@1          & R@5          & R@10         &                     \\\hline
\multirow{7}{*}{\rotatebox{90}{Flickr30k$_{EN}$}} & CLIP         &    65.0          &     87.1         &      92.2        &    85.1          &      97.3        &    \textbf{99.2}          &  87.6                   \\
& Taiyi &25.3	&48.2	&59.2	&39.3	&68.1	&79.6	&53.3 \\ 
& Wukong &-	&-	&-	&-	&-	&-	&- \\
& R2D2 &-	&-	&-	&-	&-	&-	&-\\
& CN-CLIP & 49.5	&76.9	&83.8	&66.5	&91.2 	&96.0	&77.3 \\
& AltCLIP$_{T}$ &66.3	&87.8	&\textbf{92.7}	&85.9	&97.7	&99.1	&88.3 \\
& AltCLIP &\textbf{72.5}	&\textbf{91.6}	&\textbf{95.4}	&\textbf{86.0}	&\textbf{98.0}	&99.1 &	\textbf{90.4} 
               \\\hline
\multirow{7}{*}{\rotatebox{90}{Flickr30k$_{CN}$}} & CLIP         &       0      &      2.4        &     4.0         &     2.3         &   8.1           &     12.6        &    5.0                 \\
& Taiyi &53.7	&79.8	&86.6	&63.8	&90.5	&95.9	&78.4 \\
& Wukong$^{\dag}$ &51.7	&78.9	&86.3	&76.1	&94.8	&97.5	&80.9 \\
& R2D2$^{\dag}$ &60.9	&86.8	&92.7	&77.6	&96.7	&\textbf{98.9}	&85.6 \\
& CN-CLIP &68	&89.7	&94.4	&80.2	&96.6	&98.2	&87.9 \\
& AltCLIP$_{T}$&63.7	&86.3	&92.1	&84.7	&97.4	&98.7	&87.2 \\
& AltCLIP &\textbf{69.8}	&\textbf{89.9}	&\textbf{94.7}	&\textbf{84.8}	&\textbf{97.4}	&98.8	&\textbf{89.2}   
\\\hline

\multirow{7}{*}{\rotatebox{90}{MSCOCO$_{EN}$}} & CLIP
               &36.5 &61.1  &71.1 &56.4 &79.5  &86.5  &65.2\\
&Taiyi &11.7 &27.8  &37.4 &19.8 &42.1  &54.3  &32.2 \\ 
&Wukong &-	&-	&-	&-	&-	&-	&- \\
&R2D2 &-	&-	&-	&-	&-	&-	&-\\
  
&CN-CLIP  &26.1 &50.0  &61.3 &40.9 &65.8  &76.3  &53.4\\
&AltCLIP$_{T}$&37.5 &62.1  &72.1 &57.2 &80.5  &87.5  &66.1  \\
&AltCLIP&\textbf{42.9} &\textbf{68.0}  &\textbf{77.4} &\textbf{58.6} &\textbf{80.6}  &\textbf{87.8}  &\textbf{69.2} 

               \\\hline
\multirow{7}{*}{\rotatebox{90}{MSCOCO$_{CN}(1K)$}} & CLIP  &0.6 &4.1  &7.1 &1.8 &6.7  &11.9  &5.4 \\
&Taiyi&52.0 &80.2  &89.6 &46.6 &76.3  &88.6  &72.2  \\
&Wukong$^{\dag}$ &55.2& 81.0 &90.6 &53.4 &80.2 &90.1 &75.1 \\
&R2D2$^{\dag}$ &63.3 &\textbf{89.3} &\textbf{95.7} &56.4 &85.0 &93.1 &80.5 \\
&CN-CLIP &63.7 &88.7  &94.4 &61.0 &84.7  &93.6  &81.0 \\

&AltCLIP$_{T}$&55.7 &82.9  &91.2 &60.7 &86.3  &93.4  &78.4  \\
&AltCLIP&\textbf{63.9} &87.2  &93.9 &\textbf{62.8} &\textbf{88.8}  &\textbf{95.5}  &\textbf{82.0} 

\\\hline
\multirow{7}{*}{\rotatebox{90}{MSCOCO$_{CN}(5K)$}} & CLIP  &0.8 &3.9  &5.8 &3.5 &8.9  &14.4  &6.2    \\
& Taiyi &46.1 &74.9  &85.1 &58.1 &83.9  &91.7  &73.3   \\
&Wukong &-	&-	&-	&-	&-	&-	&- \\
&R2D2&-	&-	&-	&-	&-	&-	&- \\
&CN-CLIP &58.6 &85.3  &92.7 &72.1 &90.9  &94.7  &82.4 \\
&AltCLIP$_{T}$ &54.9 &81.8  &90.5 &76.3 &93.9  &\textbf{97.6}  &82.5  \\
&AltCLIP&\textbf{61.3} &\textbf{86.0}  &\textbf{93.2} &\textbf{77.8} &\textbf{94.4}  &97.5  &\textbf{85.0} 
\\\hline
\end{tabular}
\caption{Experimental results on retrieval tasks, namely English and Chinese version of Flickr30k and MSCOCO. All image encoders used in these models are ViT-L for fair comparison. AltCLIP$_{T}$ denotes our model after Teacher Learning stage while AltCLIP denotes our model after contrastive learning stage.$^{\dag}$ represents we report original results from papers.}
\label{tab:retrieval}
\end{table*}

\subsection{Implementation details}
We initialize our text encoder from XLM-R$_{Large}$. We use the text encoder from CLIP$_{ViT-L14}$ as the teacher text encoder, and the image encoder of the same CLIP model as our image encoder. 
For specific hyper-parameter settings in our two-stage training, please refer to the table \ref{tab:hyper} in Appendix.

\begin{figure*}[htp]
    \centering
    \includegraphics[scale=0.75]{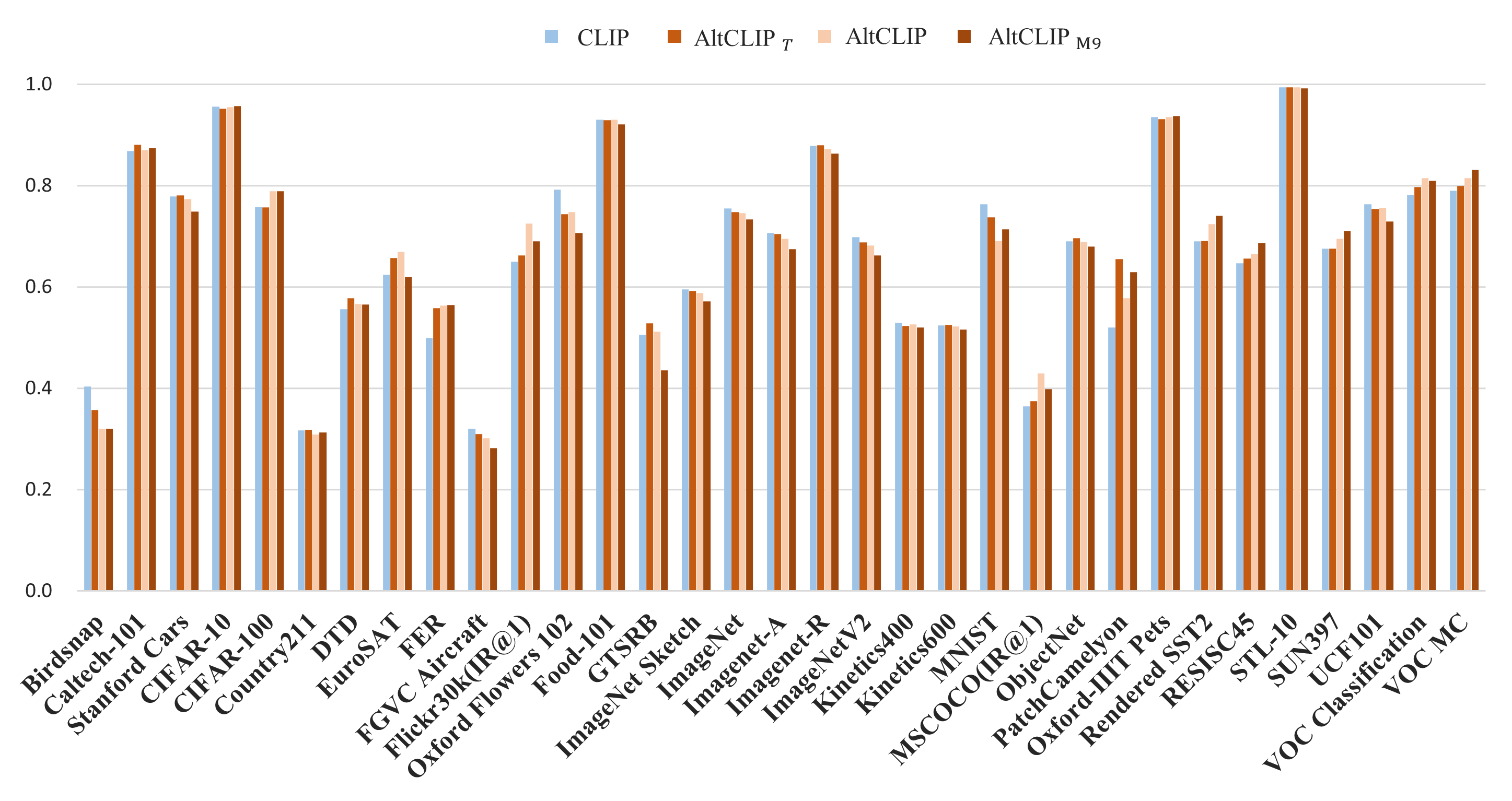}
    \caption{Comparison Results of our proposed model and OpenAI's CLIP. AltCLIP$_{T}$ denotes for our model after Teacher Learning Stage while AltCLIP denotes for our model after Contrastive Learning Stage. AltCLIP$_{M9}$ denotes for our model extending to 9 different languages. All image encoders are CLIP$_{ViT-L14}$. }
    \label{fig:clipbcm}
\end{figure*}

%% file: 5_experiments.tex
\section{Experiments}
\label{sec5}
We present our experimental results in this section. In Section \ref{sec5-1}, we introduce the datasets and metrics used in the evaluation. We comprehensively validate our model through zero-shot benchmarks in both English and Chinese in Section \ref{sec5-2}.
In Section \ref{sec5-3}, we conduct an ablation study on the effects of various design choices in Teacher Learning and Contrastive Learning. 
Finally, in Section \ref{sec5-4}, we use an AltCLIP-guided diffusion model to generate images from both Chinese and English prompts, and show that our model is capable to align text in different languages.


\subsection{Evaluation Datasets and Metrics} 
\label{sec5-1}
In this section, we describe the datasets and metrics we used in our experiments. We use ImageNet ~\cite{deng2009imagenet} and its four out-of-distribution test variants, i.e. ImageNet Sketch~\cite{wang2019learning}, ImageNet-A~\cite{hendrycks2021natural}, ImageNet-R~\cite{hendrycks2021many}, ImageNetV2~\cite{recht2019imagenet}, to evaluate zero-shot image classification performances in English and Chinese. \footnote{We use the Chinese translation of classnames from \url{https://github.com/ningbonb/imagenet_classes_chinese}}
During evaluation, we adapt the templates of manual prompts from CLIP for English and corresponding machine translation templates for Chinese. For cross-modal retrieval, we select Flickr30k~\cite{young2014image}, MSCOCO~\cite{lin2014microsoft}, as well as their corresponding Chinese datasets, Flickr30k$_\text{CN}$~\cite{lan2017fluency}, MSCOCO$_\text{CN}$~ \footnote{two versions, texts in 1k version are manual written captions while in 5k version are manual translated captions}~\cite{li2019coco}, to evaluate zero-shot image-to-text retrieval and text-to-image retrieval performances. 

We further validate our model on a wide range of tasks to compare the performance with the original CLIP model. We collect the datasets introduced in CLIP and the Open CLIP benchmark \footnote{\url{https://github.com/LAION-AI/CLIP_benchmark}}, including Birdsnap~\cite{berg2014birdsnap}, Caltech-101~\cite{fei2006one}, Stanford Cars~\cite{Krause2013}, CIFAR-10~\cite{krizhevsky2009learning}, CIFAR-100~\cite{krizhevsky2009learning}, Country211~\cite{radford2021learning}, DTD~\cite{cimpoi2014describing}, EuroSAT~\cite{helber2019eurosat}, Facial Emotion Recognition 2013~\cite{goodfellow2013challenges}, FGVC Aircraft~\cite{blaschko2012towards}, Oxford Flowers 102~\cite{nilsback2008automated}, Food-101~\cite{bossard14}, GTSRB~\cite{stallkamp2011german}, Kinetics400~\cite{kay2017kinetics}, Kinetics600~\cite{carreira2018short}, MNIST~\cite{cirecsan2011high}, PatchCamelyon~\cite{veeling2018rotation}, ObjectNet~\cite{barbu2019objectnet}, Oxford-IIIT Pets~\cite{parkhi2012cats}, Rendered SST2~\cite{radford2021learning}, RESISC45~\cite{cheng2017remote}, STL-10~\cite{coates2011analysis}, SUN397~\cite{Xiao2010}, UCF101~\cite{soomro2012ucf101}, Pascal VOC 2007 Classification~\cite{pascal-voc-2007}, Pascal VOC 2007 Multilabel Classification~\cite{pascal-voc-2007}. 
Finally, we evaluate our multilingual model AltCLIP$_{M9}$ on the \textbf{XTD} ~\cite{aggarwal2020zeroshot} dataset. XTD selected 1K images from MS COCO, and translated the corresponding English Captions into 11 languages (English(en), German(de), French(fr), Chinese(zh), Japanese(ja), Italian(it), Spanish(es), Russian(ru), Polish(pl), Turkish(tr), Korean(ko)).
The evaluation metrics for image classification benchmarks are respectively accuracy (default), mean per class (the average of recall obtained on each class, for imbalanced datasets, i.e. FGVC Aircraft, Oxford-IIIT Pets, Caltech-101, Oxford Flowers 102), 11-point mAP (mean average of 11-pt interpolated precision for each class, for VOC 2007), mean(top1, top5) (the mean of acc@1 and acc@5, for Kinetics400 and Kinetics600), while for cross-modal retrieval benchmarks, are Recall@K where $K\in\{1,5,10\}$, and Mean Recall(MR, i.e., the average of
Recall@K) for both image-to-text retrieval and text-to-image retrieval tasks, which are same with the setups in CLIP ~\cite{radford2021learning}.

\begin{table*}[t]
\centering
\small
\begin{tabular}{ccccccccc} 
\hline
\diagbox{Method}{Language} & En            & Es            & Fr            & Zh            & It            & Ko            & Ru            & Jp             \\ 
\hline
CLIP$_{ViT-B32}$                       & 90.3          & -             & -             & -             & -             & -             & -             & -              \\
CLIP$_{ViT-L14}$                       & 91.8          & -             & -             & -             & -             & -             & -             & -              \\
CLIP$_{ViT-B16+}$$\dag$                       & 94.3          & -             & -             & -             & -             & -             & -             & -              \\
mUSE PATR                      &83.6  &75.6 &76.9 &76.1 &73.4 &64.3 &73.6  &69.4            \\
mUSE m3             &85.3  &78.9 &78.9 &76.7 &73.6 &67.8 &76.1  &70.7         \\
M-CLIP$_{ViT-B32}$                     &91.8		&89.1	&89.4	&89.3	&89.8		&82.1	&86.1		&81.0           \\
M-CLIP$_{ViT-L14}$                     & 92.4          & 91            & 90            & 89.7          & 91.1          & 85.2          & 85.8          & 81.9           \\
M-CLIP$_{ViT-B16+}$$\dag$      &95   &93.6	&\textbf{93.1}	&94	&93.1		&89	&90	&84.2      \\
AltCLIP$_{M9}$                    & \textbf{95.4} & \textbf{94.1} & 92.9 & \textbf{95.1} & \textbf{94.2} & \textbf{94.4} & \textbf{91.8} & \textbf{91.7}  \\
\hline
\end{tabular}
\caption{Comparison results on the multilingual cross-modal Retrieve dataset XTD.  Following M-CLIP, we report recall@10 on Image to Text. \dag  represents using image encoder released by OpenCLIP project ~\cite{ilharco_gabriel_2021_5143773}. 
\label{tab:xtd}}
\end{table*}

\begin{table*}[tb]
\centering
\small
\begin{tabular}{cccc|cccc} 
\toprule
EN-EN & EN-CN$_{MT}$ & EN-CN$_{HT}$ & CL             & \multicolumn{1}{l}{Flickr30K$_{EN}$} & \multicolumn{1}{l}{Flickr30K$_{CN}$} & \multicolumn{1}{l}{ImageNet$_{EN}$} & \multicolumn{1}{l}{ImageNet$_{CN}$}  \\ 
\midrule
\Checkmark                &\Checkmark&\Checkmark&\Checkmark& \textbf{90.4}                                 & \textbf{89.2}                                   & 74.5                                & \textbf{59.6}                           \\

\Checkmark             &\Checkmark&\Checkmark&& 88.3                                 & 87.2                                   & \textbf{74.7}                                & 58.2                                 \\

\Checkmark &\Checkmark&&& 86.8                                & 85.8                                  & 51.6                               & 41.7                                \\

\Checkmark  &&&& 86.6                                & 53.9                                  & 53.8                               & 12.8                                \\

   &\Checkmark&&& 61.9                               & 85.4                                  & 15.5                               & 42.5                                \\
\bottomrule
\end{tabular}
\caption{Ablation Experiments. CL indicates the use of Contrastive Learning stage, while EN-EN, \text{EN-CN}$_{MT}$, \text{EN-CN}$_{HT}$ refers to parallel data used in Teacher Learning stage. Specifically, EN-EN indicates the use of English-English text pairs; EN-CN indicates the use of English-Chinese parallel text, including \text{EN-CN}$_{MT}$ represents machine translated pairs while \text{EN-CN}$_{HT}$ stands for human-translated data, i.e TSL2019. All compared models are pre-trained for 10 epochs. $\label{tab:pkd}$}
\end{table*}

\subsection{Zero-shot performance}
\label{sec5-2}
\paragraph{Image Classification}
We first present evaluation results of zero-shot image classification on the ImageNet dataset and its four out-of-distribution variants. For baselines, we compare our model with CLIP ~\cite{radford2021learning}, CN-CLIP ~\cite{yang2022chinese}, and multilingual CLIP (M-CLIP) ~\cite{carlsson2022cross}.
As illustrated in Table~\ref{tab:imagenet}, AltCLIP achieves results that are very close to CLIP in English and sets new state-of-the-art results on Chinese ImageNet, ImageNet-A, ImageNet-R, and ImageNet V2. These results demonstrate the effectiveness of our method: compared to other Chinese baseline models where hundreds of millions of text-image pairs are used in pretraining, we only use 36M parallel text data and 2M text-image pairs in training.

\paragraph{Cross-modal Retrieval}
For cross-modal retrieval, we compare our model with CLIP in English and R2D2 ~\cite{xie2022zero}, Wukong ~\cite{gu2022wukong}, Taiyi ~\cite{fengshenbang} and CN-CLIP ~\cite{yang2022chinese} in Chinese. The results are shown in Table~\ref{tab:retrieval}. AltCLIP outperforms all baseline models on most datasets and tasks. We notice that AltCLIP outperforms CLIP on both text-to-image and image-to-text retrieval. This could be due to the following reasons: 1). We used a small subset (less than 1M) of LAION 5B at the Contrastive Learning stage, which is in a different distribution of the pretraining data used in CLIP; 2). Our language encoder initialized from XLM-R provides better language understanding ability. We left it as a future investigation to properly analyze the factors.

\paragraph{Multilingual Cross-modal Retrieval}
For baseline, we compare our model with CLIP, M-CLIP ~\cite{carlsson2022cross} and mUSE ~\cite{yang-etal-2020-multilingual}. The results of the comparison on XTD ~\cite{aggarwal2020zeroshot} are shown in Table \ref{tab:xtd}, where AltCLIP$_{M9}$ achieves state-of-the-art results in 7 languages. Notably, joint multilingual training maintains state-of-the-art capabilities in English and achieves better results than the original CLIP model on this dataset. We may achieve better results for two reasons: 1. We equally translate a fixed set of English captions for each language, which allows each language to share a more comprehensive distribution. 2. We introduce more parallel data of human translation, and experiments show that this part of the data is helpful for establishing a robust multilingual representation.

\paragraph{Full CLIP benchmark}
We present the evaluation results of a wider range of tasks in English in Figure \ref{fig:clipbcm}. We compare the effectiveness of the Teacher Learning Stage and Contrastive Learning Stage, and the bilingual AltCLIP and multilingual AltCLIP$_{M9}$. We observe that at the Teacher Learning Stage, the model already learns a good representation of text-image representation, as it gets a close performance of the original CLIP model on a broad range of zero-shot benchmarks. The Contrastive Learning stage further improves our model’s performance, especially on retrieval tasks like Flickr30k and MSCOCO. Further, the multilingual AltCLIP$_{M9}$ achieves close performances with bilingual AltCLIP, indicating that our method is effective at extending the number of languages supported.

\subsection{Ablation study}
\label{sec5-3}
In this section, we present results from ablation studies. We show the significance of including various parallel data in Teacher Learning stage in Table \ref{tab:pkd}. As illustrated in the 3rd and 5th line, without English-to-English parallel data, the accuracy on English ImageNet drastically drops to 15.47 from 53.8. Similarly, excluding machine-translated English-to-Chinese data, has a great impact on the performances on Chinese benchmarks, i.e. Imagenet$_{CN}$ and Flickr30K$_{CN}$, due to influenced Chinese text-image representation. Moreover, empirical experiments show that introducing human translated parallel data leads to a great improvement on Imagenet$_{CN}$ which may be related to the distribution of the data set. This indicates that the diversity of training data used for teacher learning, can benefit the language model to gain more knowledge about entities or concepts.


\subsection{Examples of text-to-image generation}
\label{sec5-4}
To demonstrate the effects of text alignment in our model, we use the text encoder of AltCLIP and AltCLIP$_{M9}$ to further finetune a text-to-image generation model from Stable Diffusion ~\cite{rombach2022high}.
We use stable-diffusion v1-4\footnote{https://huggingface.co/CompVis/stable-diffusion-v-1-4-original} as initialization, and we use AltCLIP or AltCLIP$_{M9}$ as the language encoder. We freeze all the parameters in the diffusion model except for the key and value projection layers of the cross-attention block during finetuning. The dataset we used to finetune is the same one used for Contrastive Learning as described in Section \ref{training data}. We call our image generation models AltDiffusion and AltDiffusion$_{M9}$. 

\begin{figure}
    \centering
    \subfigure[Stable Diffusion]{
    \includegraphics[width=0.45\textwidth]{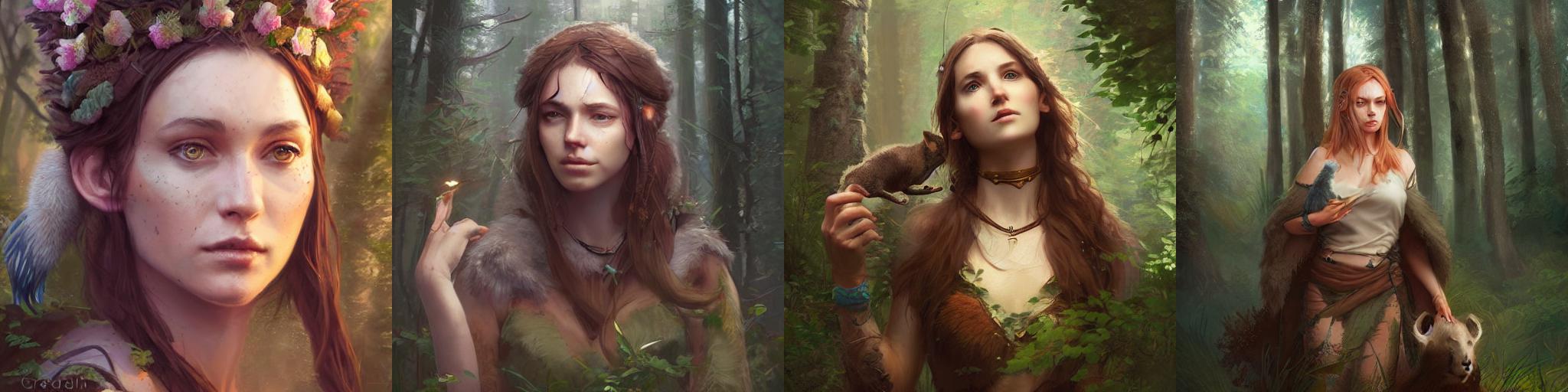}
    }
    \subfigure[AltDiffusion EN]{
    \includegraphics[width=0.45\textwidth]{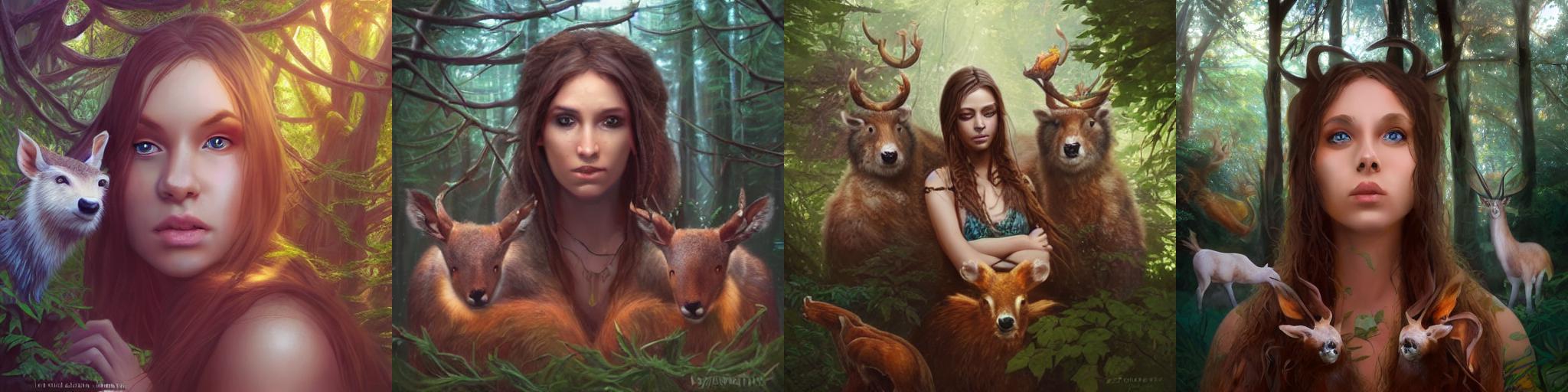}
    }
    \subfigure[AltDiffusion CN]{
    \includegraphics[width=0.45\textwidth]{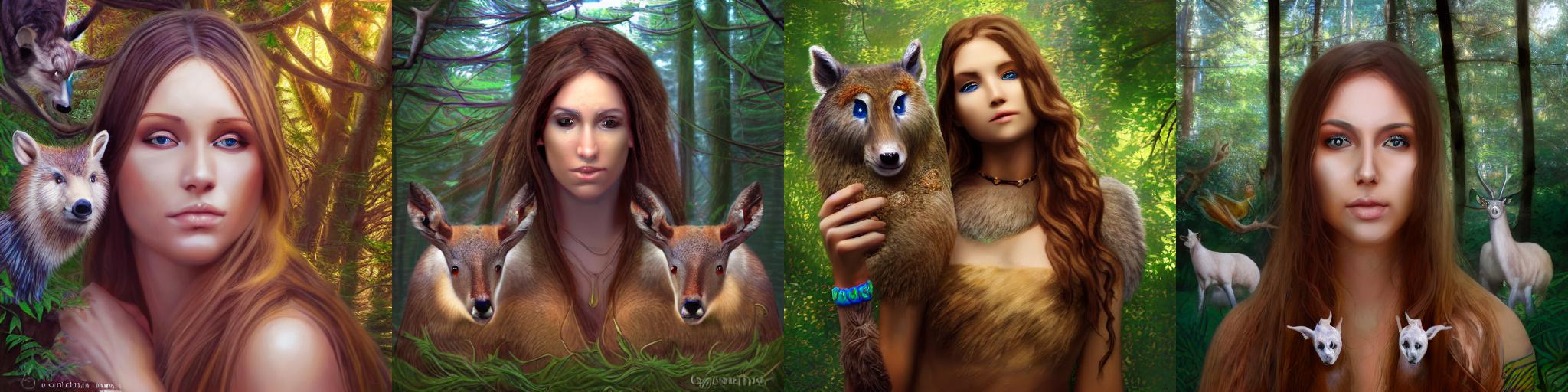}
    }
    \caption{Examples of text-to-image generation. Text prompt: \textit{"a pretty female druid surrounded by forest animals, digital painting, photorealistic, in the style of greg rutkowski, highly detailed, realistic.",\begin{CJK}{UTF8}{gbsn} "一个由森林动物环绕的漂亮的女德鲁伊,数字绘画,摄影现实,格雷格·鲁特科夫斯基风格,高度详细,现实"\end{CJK}} \label{fig:demo}}
    \label{fig:my_label}
\end{figure}

As shown in Fig.~\ref{fig:demo}, AltDiffusion generates high-quality images close to Stable Diffusion on both English and Chinese prompts. Further, we notice that our model generates very similar images given translated English and Chinese prompts. For the multilingual version, we observed a very interesting phenomenon: when given a prompt that’s translated into different languages, the generated images reflect cultural differences to some extent. See Tab. ~\ref{tab:m9} for some examples. We left it as a future effort to explore and analyze carefully on how culture inherited in different languages can affect image generation.


\begin{table*}[]
    \centering
    \begin{tabular}{b{0.25\linewidth}|p{0.65\linewidth}}\hline
        Prompts & Generated Image \\\hline
        \includegraphics[width=0.95\linewidth]{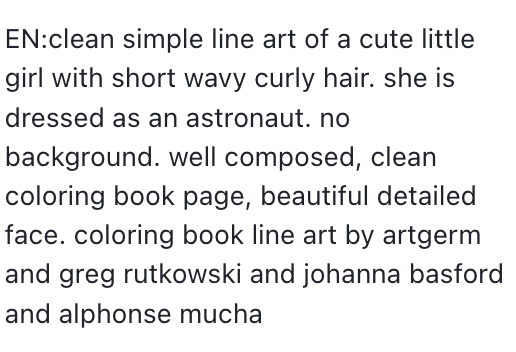}  & \includegraphics[width=1\linewidth]{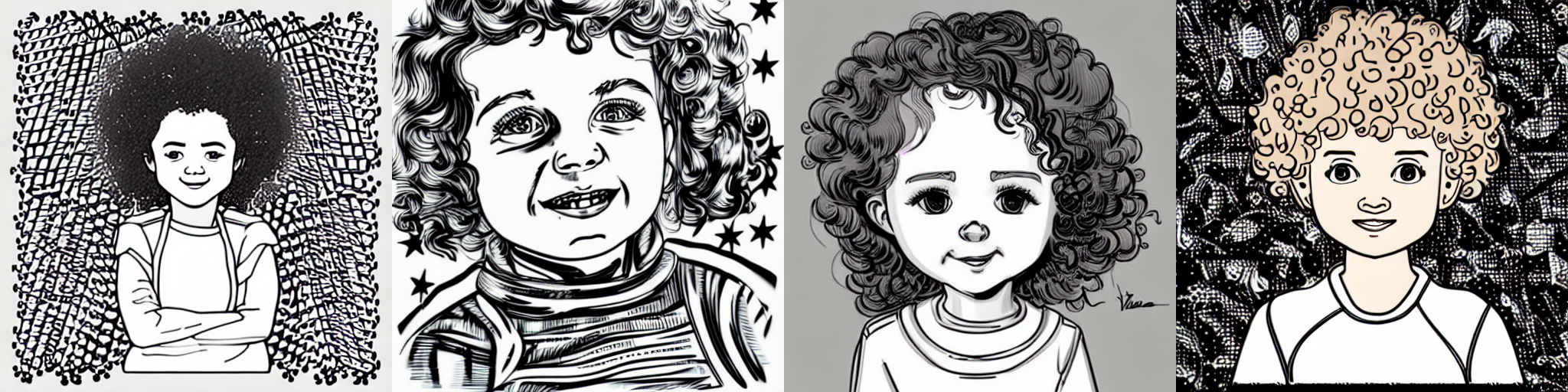}\\
      \includegraphics[width=0.95\linewidth]{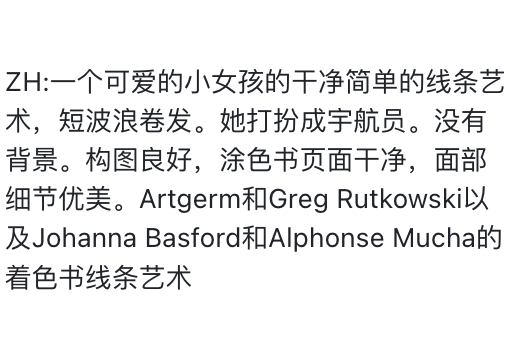}  & \includegraphics[width=1\linewidth]{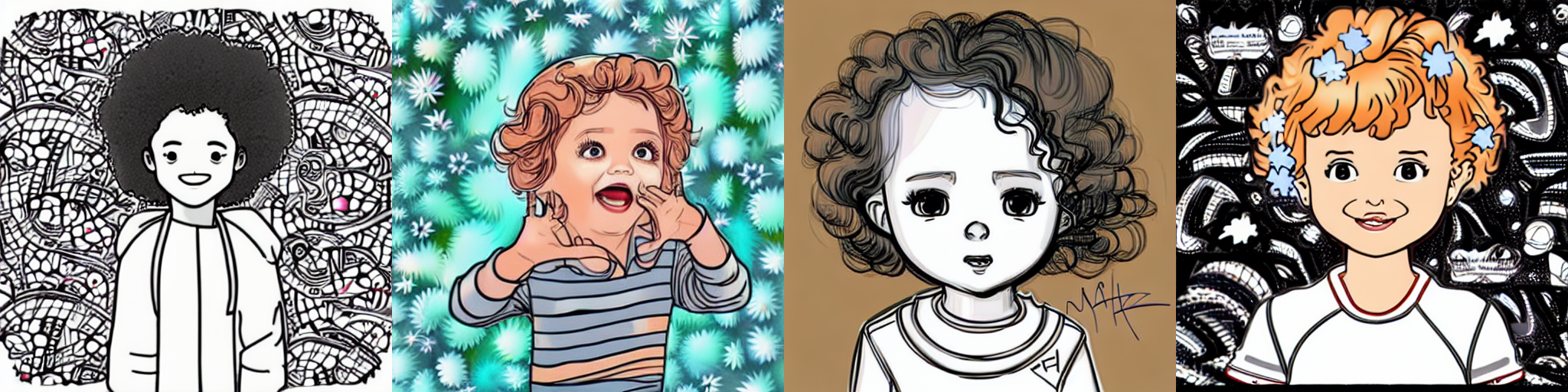}\\
      \includegraphics[width=0.95\linewidth]{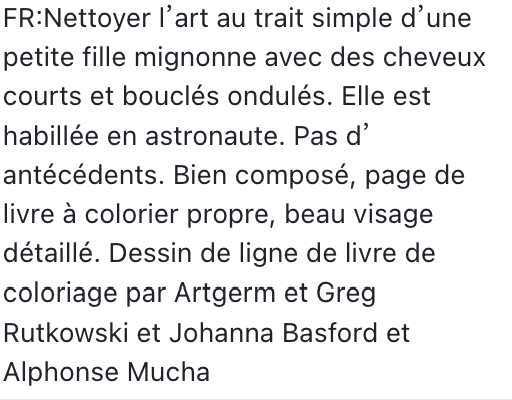}  & \includegraphics[width=1\linewidth]{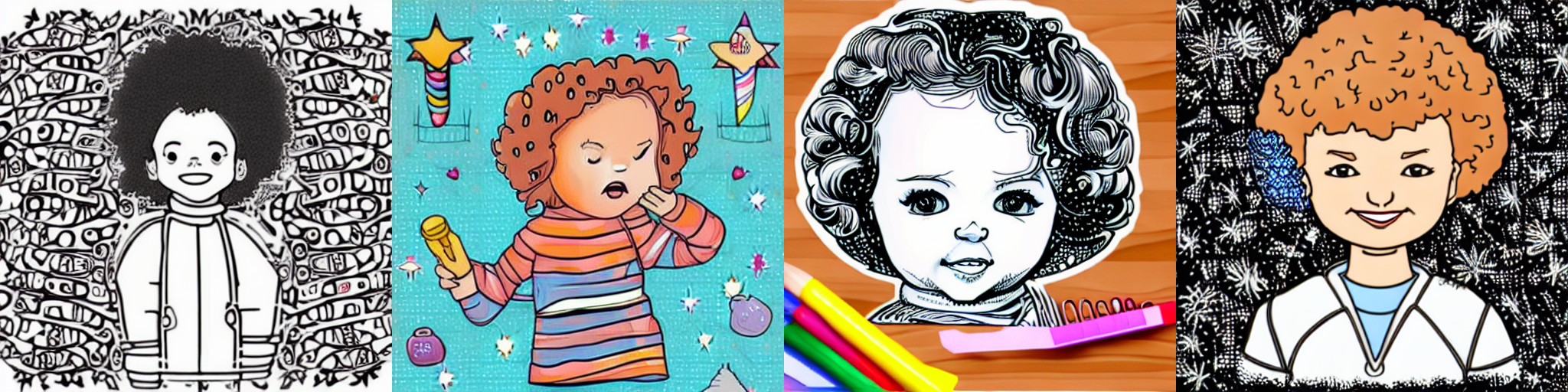}\\
      \includegraphics[width=0.95\linewidth]{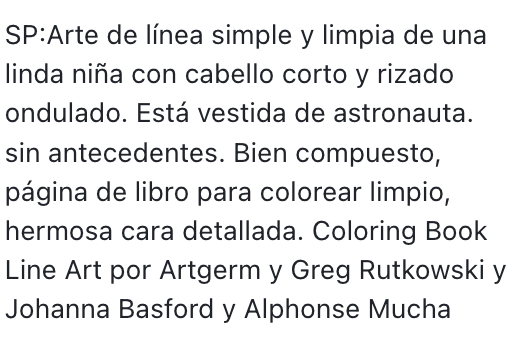}   & \includegraphics[width=1\linewidth]{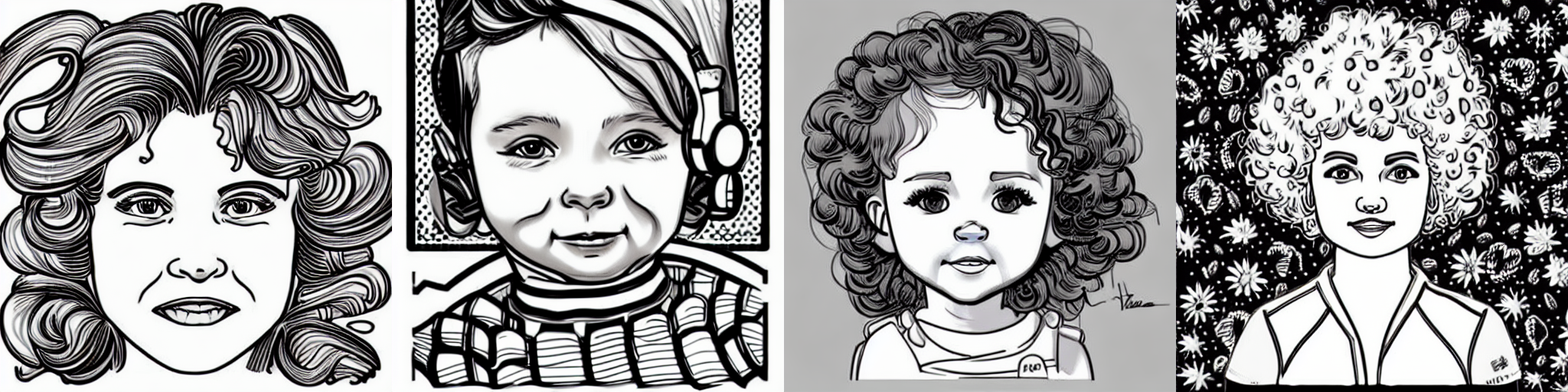}\\
       \includegraphics[width=0.95\linewidth]{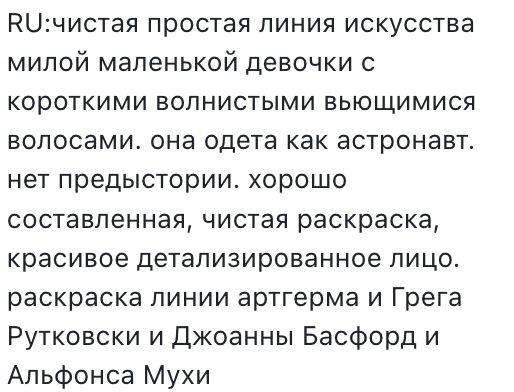}  & \includegraphics[width=1\linewidth]{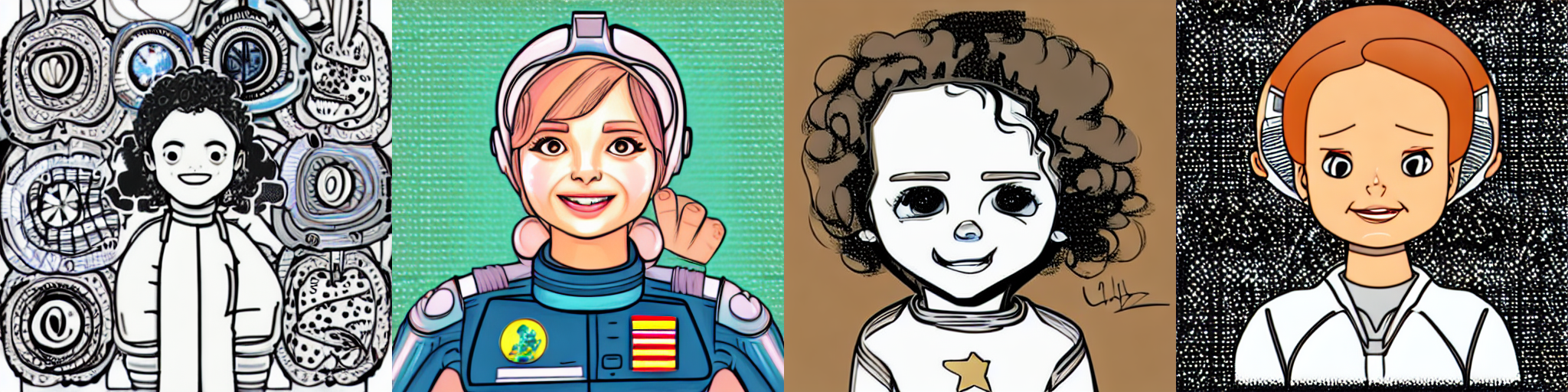}\\
        \includegraphics[width=0.95\linewidth]{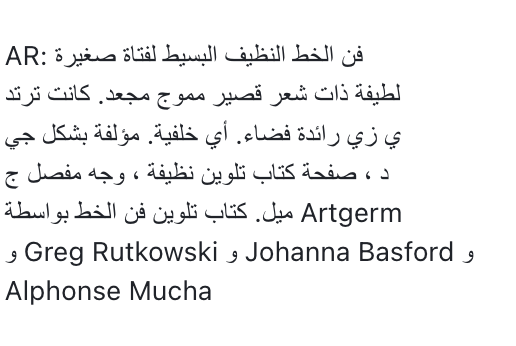}  & \includegraphics[width=1\linewidth]{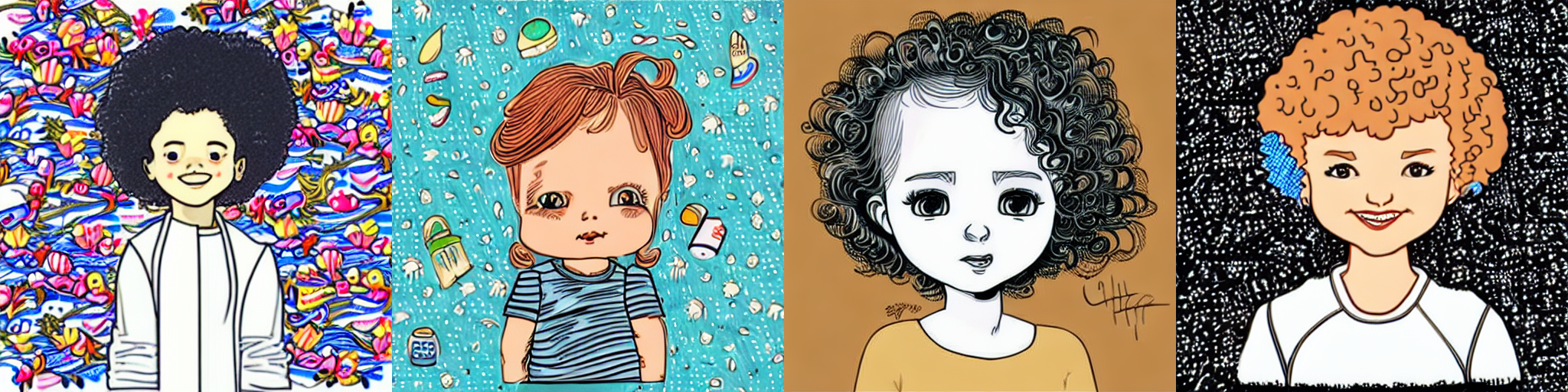}\\
         \includegraphics[width=0.95\linewidth]{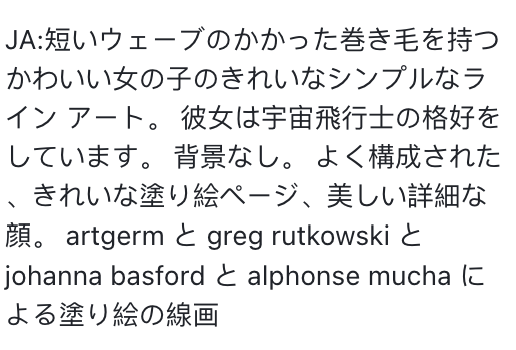}   & \includegraphics[width=1\linewidth]{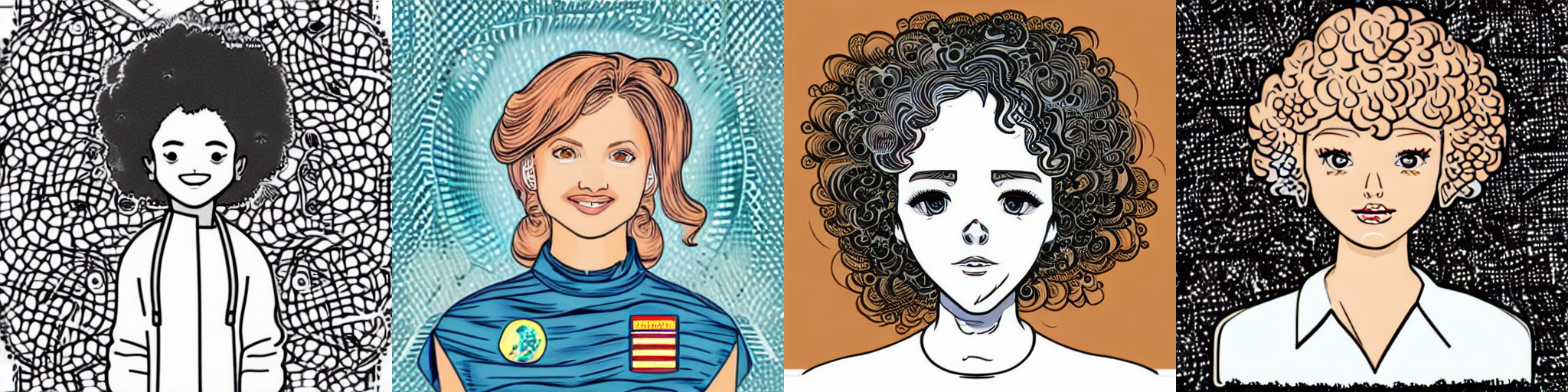}\\
          \includegraphics[width=0.95\linewidth]{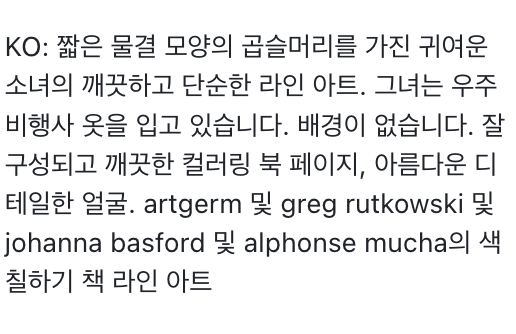}   & \includegraphics[width=1\linewidth]{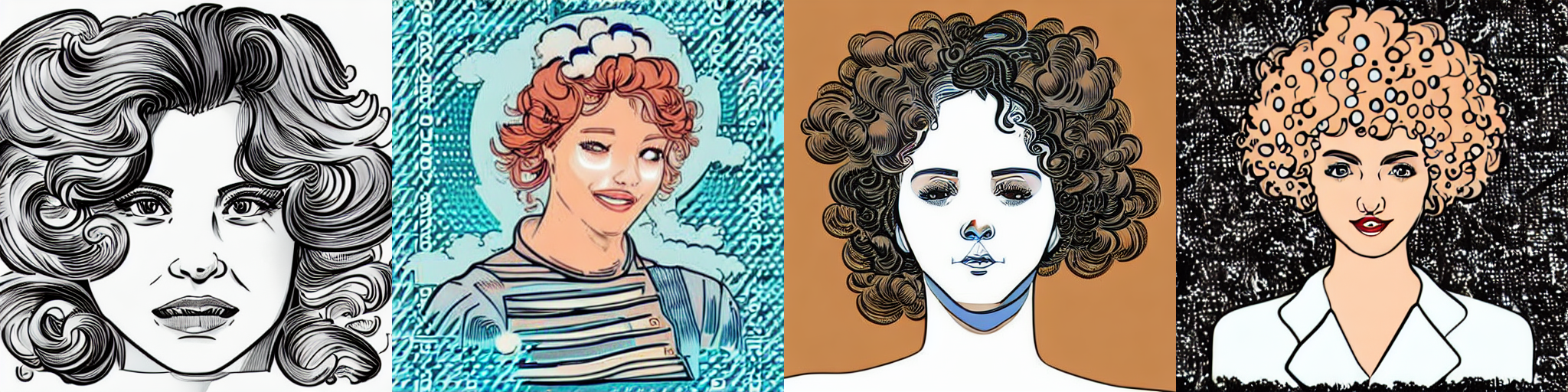}\\
           \includegraphics[width=0.95\linewidth]{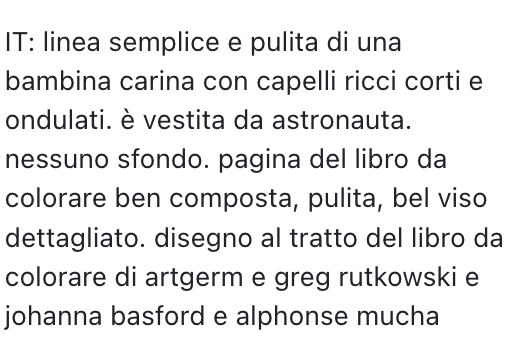}   & \includegraphics[width=1\linewidth]{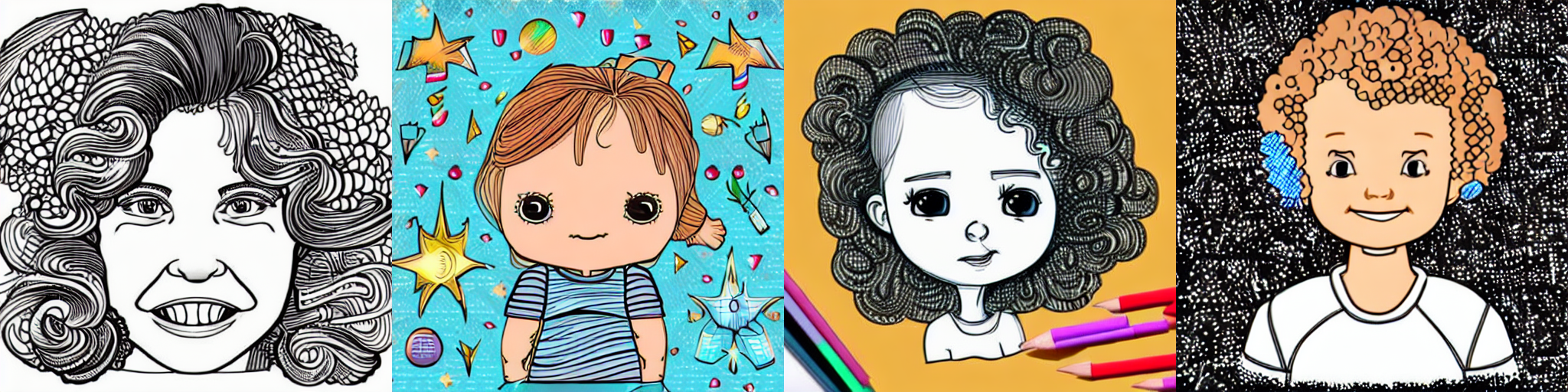}\\\hline
    \end{tabular}
    \caption{The images generated by AltDiffusion$_{M9}$ with the same prompt translated to nine languages and a fixed seed.}
    \label{tab:m9}
\end{table*}

%% file: 6_conclusion.tex
\section{Conclusion}
In this work, we presented a conceptually simple and effective two-stage training schema to learn bilingual and multilingual multimodal representation models, via teacher learning and contrastive learning. We show the effectiveness of our method by conducting extensive experiments on a wide range of tasks in English and Chinese. In Chinese, our method sets new state-of-the-art results on multiple zero-shot image classification and retrieval tasks while being highly effective: we only use tens of millions of text data and two million text-image pairs during training, whereas most prior work requires training on hundreds of millions of text-image pairs. Future work includes exploring altering the image encoder to combine vision signals learned from different data distributions and possibly eliminating the need for machine-translated data to build a multilingual multimodal pretraining model.

%% file: 7_acknowledgement.tex
\section*{Acknowledgements}
We’d like to thank Yizhou Zheng for helpful suggestions on fine-tuning stable diffusion models. We’d like to thank Luke Zettlemoyer for valuable suggestions and advices. We also thank Chenghua Zhou, Quan Sun and Yue Cao for helpful discussions and contributions. Finally, we’d like to thank the data and infrastructure team at BAAI for their support on this project.

%% file: A_appendix.tex
\section{Hyper-parameters Optimization}
\begin{table}[htb]
\small
\centering
\begin{tabular}{l|cc} 
\toprule
\multicolumn{1}{l}{\small{Hyper-paramters}} & \small{Teacher Learning}                             & \small{Contrastive Learning }                                  \\ 
\hline\hline
Batch size                          & 1024                                         & 1024                                          \\
Optimizer~(AdamW, $\beta$)  & (0.99, 0.999) & (0.99, 0.999)  \\
Learning rate                       & 1e-4                                         & 2e-6                                          \\
Weight decay                        & 1e-1                                         & 5e-2                                          \\
Eps                                 & 1e-8                                         & 1e-8                                          \\
Warmup steps                        & 500                                          & 2000                                          \\
\#Epochs                             & 10                                           & 1                                             \\
Gradient clipping                   & 1.0                                          & 5.0                                           \\
Steps                               & 238620                                       & 2000                                             \\
\bottomrule
\end{tabular}
\caption{Hyper-parameters setting in Teacher Learning Stage and Constrastive Learning Stage.\label{tab:ccl}}
\label{tab:hyper}
\end{table}

%% file: main.bbl
\begin{thebibliography}{60}
\expandafter\ifx\csname natexlab\endcsname\relax\def\natexlab#1{#1}\fi

\bibitem[{Aggarwal and Kale(2020{\natexlab{a}})}]{aggarwal2020towards}
Pranav Aggarwal and Ajinkya Kale. 2020{\natexlab{a}}.
\newblock Towards zero-shot cross-lingual image retrieval.
\newblock \emph{arXiv preprint arXiv:2012.05107}.

\bibitem[{Aggarwal and Kale(2020{\natexlab{b}})}]{aggarwal2020zeroshot}
Pranav Aggarwal and Ajinkya Kale. 2020{\natexlab{b}}.
\newblock \href {http://arxiv.org/abs/2012.05107} {Towards zero-shot
  cross-lingual image retrieval}.

\bibitem[{Barbu et~al.(2019)Barbu, Mayo, Alverio, Luo, Wang, Gutfreund,
  Tenenbaum, and Katz}]{barbu2019objectnet}
Andrei Barbu, David Mayo, Julian Alverio, William Luo, Christopher Wang, Dan
  Gutfreund, Josh Tenenbaum, and Boris Katz. 2019.
\newblock Objectnet: A large-scale bias-controlled dataset for pushing the
  limits of object recognition models.
\newblock \emph{Advances in neural information processing systems}, 32.

\bibitem[{Berg et~al.(2014)Berg, Liu, Woo~Lee, Alexander, Jacobs, and
  Belhumeur}]{berg2014birdsnap}
Thomas Berg, Jiongxin Liu, Seung Woo~Lee, Michelle~L Alexander, David~W Jacobs,
  and Peter~N Belhumeur. 2014.
\newblock Birdsnap: Large-scale fine-grained visual categorization of birds.
\newblock In \emph{Proceedings of the IEEE Conference on Computer Vision and
  Pattern Recognition}, pages 2011--2018.

\bibitem[{Bianchi et~al.(2021)Bianchi, Attanasio, Pisoni, Terragni, Sarti, and
  Lakshmi}]{bianchi2021contrastive}
Federico Bianchi, Giuseppe Attanasio, Raphael Pisoni, Silvia Terragni, Gabriele
  Sarti, and Sri Lakshmi. 2021.
\newblock Contrastive language-image pre-training for the italian language.
\newblock \emph{arXiv preprint arXiv:2108.08688}.

\bibitem[{Blaschko et~al.(2012)Blaschko, Girshick, Kannala, Kokkinos,
  Mahendran, Maji, Mohammed, Rahtu, Saphra, Simonyan
  et~al.}]{blaschko2012towards}
Matthew Blaschko, Ross~B Girshick, Juho Kannala, Iasonas Kokkinos, Siddarth
  Mahendran, Subhransu Maji, Sammy Mohammed, Esa Rahtu, Naomi Saphra, Karen
  Simonyan, et~al. 2012.
\newblock Towards a detailed understanding of objects and scenes in natural
  images.

\bibitem[{Bossard et~al.(2014)Bossard, Guillaumin, and Van~Gool}]{bossard14}
Lukas Bossard, Matthieu Guillaumin, and Luc Van~Gool. 2014.
\newblock Food-101 -- mining discriminative components with random forests.
\newblock In \emph{European Conference on Computer Vision}.

\bibitem[{Carlsson et~al.(2022)Carlsson, Eisen, Rekathati, and
  Sahlgren}]{carlsson2022cross}
Fredrik Carlsson, Philipp Eisen, Faton Rekathati, and Magnus Sahlgren. 2022.
\newblock Cross-lingual and multilingual clip.
\newblock In \emph{Proceedings of the Thirteenth Language Resources and
  Evaluation Conference}, pages 6848--6854.

\bibitem[{Carreira et~al.(2018)Carreira, Noland, Banki-Horvath, Hillier, and
  Zisserman}]{carreira2018short}
Joao Carreira, Eric Noland, Andras Banki-Horvath, Chloe Hillier, and Andrew
  Zisserman. 2018.
\newblock A short note about kinetics-600.
\newblock \emph{arXiv preprint arXiv:1808.01340}.

\bibitem[{Changpinyo et~al.(2021)Changpinyo, Sharma, Ding, and
  Soricut}]{changpinyo2021conceptual}
Soravit Changpinyo, Piyush Sharma, Nan Ding, and Radu Soricut. 2021.
\newblock Conceptual 12m: Pushing web-scale image-text pre-training to
  recognize long-tail visual concepts.
\newblock In \emph{Proceedings of the IEEE/CVF Conference on Computer Vision
  and Pattern Recognition}, pages 3558--3568.

\bibitem[{Chen et~al.(2015)Chen, Fang, Lin, Vedantam, Gupta, Doll{\'a}r, and
  Zitnick}]{chen2015microsoft}
Xinlei Chen, Hao Fang, Tsung-Yi Lin, Ramakrishna Vedantam, Saurabh Gupta, Piotr
  Doll{\'a}r, and C~Lawrence Zitnick. 2015.
\newblock Microsoft coco captions: Data collection and evaluation server.
\newblock \emph{arXiv preprint arXiv:1504.00325}.

\bibitem[{Cheng et~al.(2017)Cheng, Han, and Lu}]{cheng2017remote}
Gong Cheng, Junwei Han, and Xiaoqiang Lu. 2017.
\newblock Remote sensing image scene classification: Benchmark and state of the
  art.
\newblock \emph{Proceedings of the IEEE}, 105(10):1865--1883.

\bibitem[{Cimpoi et~al.(2014)Cimpoi, Maji, Kokkinos, Mohamed, and
  Vedaldi}]{cimpoi2014describing}
Mircea Cimpoi, Subhransu Maji, Iasonas Kokkinos, Sammy Mohamed, and Andrea
  Vedaldi. 2014.
\newblock Describing textures in the wild.
\newblock In \emph{Proceedings of the IEEE conference on computer vision and
  pattern recognition}, pages 3606--3613.

\bibitem[{Cire{\c{s}}an et~al.(2011)Cire{\c{s}}an, Meier, Masci, Gambardella,
  and Schmidhuber}]{cirecsan2011high}
Dan~C Cire{\c{s}}an, Ueli Meier, Jonathan Masci, Luca~M Gambardella, and
  J{\"u}rgen Schmidhuber. 2011.
\newblock High-performance neural networks for visual object classification.
\newblock \emph{arXiv preprint arXiv:1102.0183}.

\bibitem[{Coates et~al.(2011)Coates, Ng, and Lee}]{coates2011analysis}
Adam Coates, Andrew Ng, and Honglak Lee. 2011.
\newblock An analysis of single-layer networks in unsupervised feature
  learning.
\newblock In \emph{Proceedings of the fourteenth international conference on
  artificial intelligence and statistics}, pages 215--223. JMLR Workshop and
  Conference Proceedings.

\bibitem[{Conneau et~al.(2020)Conneau, Khandelwal, Goyal, Chaudhary, Wenzek,
  Guzm{\'a}n, Grave, Ott, Zettlemoyer, and Stoyanov}]{conneau2020unsupervised}
Alexis Conneau, Kartikay Khandelwal, Naman Goyal, Vishrav Chaudhary, Guillaume
  Wenzek, Francisco Guzm{\'a}n, Edouard Grave, Myle Ott, Luke Zettlemoyer, and
  Veselin Stoyanov. 2020.
\newblock Unsupervised cross-lingual representation learning at scale.
\newblock In \emph{ACL}.

\bibitem[{Deng et~al.(2009)Deng, Dong, Socher, Li, Li, and
  Fei-Fei}]{deng2009imagenet}
Jia Deng, Wei Dong, Richard Socher, Li-Jia Li, Kai Li, and Li~Fei-Fei. 2009.
\newblock Imagenet: A large-scale hierarchical image database.
\newblock In \emph{CVPR}, pages 248--255. Ieee.

\bibitem[{Dosovitskiy et~al.(2020)Dosovitskiy, Beyer, Kolesnikov, Weissenborn,
  Zhai, Unterthiner, Dehghani, Minderer, Heigold, Gelly
  et~al.}]{dosovitskiy2020image}
Alexey Dosovitskiy, Lucas Beyer, Alexander Kolesnikov, Dirk Weissenborn,
  Xiaohua Zhai, Thomas Unterthiner, Mostafa Dehghani, Matthias Minderer, Georg
  Heigold, Sylvain Gelly, et~al. 2020.
\newblock An image is worth 16x16 words: Transformers for image recognition at
  scale.
\newblock \emph{arXiv preprint arXiv:2010.11929}.

\bibitem[{Everingham(2007)}]{pascal-voc-2007}
Mark Everingham. 2007.
\newblock The pascal visual object classes challenge,(voc2007) results.
\newblock \emph{http://pascallin.ecs.soton.ac.uk/challenges/VOC/
  voc2007/index.html.}

\bibitem[{Fei et~al.(2021)Fei, Lu, Gao, Yang, Huo, Wen, Lu, Song, Gao, Xiang
  et~al.}]{fei2021wenlan}
Nanyi Fei, Zhiwu Lu, Yizhao Gao, Guoxing Yang, Yuqi Huo, Jingyuan Wen, Haoyu
  Lu, Ruihua Song, Xin Gao, Tao Xiang, et~al. 2021.
\newblock Wenlan 2.0: Make ai imagine via a multimodal foundation model.
\newblock \emph{arXiv preprint arXiv:2110.14378}.

\bibitem[{Fei-Fei et~al.(2006)Fei-Fei, Fergus, and Perona}]{fei2006one}
Li~Fei-Fei, Robert Fergus, and Pietro Perona. 2006.
\newblock One-shot learning of object categories.
\newblock \emph{IEEE transactions on pattern analysis and machine
  intelligence}, 28(4):594--611.

\bibitem[{Goodfellow et~al.(2013)Goodfellow, Erhan, Carrier, Courville, Mirza,
  Hamner, Cukierski, Tang, Thaler, Lee et~al.}]{goodfellow2013challenges}
Ian~J Goodfellow, Dumitru Erhan, Pierre~Luc Carrier, Aaron Courville, Mehdi
  Mirza, Ben Hamner, Will Cukierski, Yichuan Tang, David Thaler, Dong-Hyun Lee,
  et~al. 2013.
\newblock Challenges in representation learning: A report on three machine
  learning contests.
\newblock In \emph{International conference on neural information processing},
  pages 117--124. Springer.

\bibitem[{Gu et~al.(2022)Gu, Meng, Lu, Hou, Niu, Xu, Liang, Zhang, Jiang, and
  Xu}]{gu2022wukong}
Jiaxi Gu, Xiaojun Meng, Guansong Lu, Lu~Hou, Minzhe Niu, Hang Xu, Xiaodan
  Liang, Wei Zhang, Xin Jiang, and Chunjing Xu. 2022.
\newblock Wukong: 100 million large-scale chinese cross-modal pre-training
  dataset and a foundation framework.
\newblock \emph{arXiv preprint arXiv:2202.06767}.

\bibitem[{Hadsell et~al.(2006)Hadsell, Chopra, and
  LeCun}]{hadsell2006dimensionality}
Raia Hadsell, Sumit Chopra, and Yann LeCun. 2006.
\newblock Dimensionality reduction by learning an invariant mapping.
\newblock In \emph{2006 IEEE Computer Society Conference on Computer Vision and
  Pattern Recognition (CVPR'06)}, volume~2, pages 1735--1742. IEEE.

\bibitem[{Helber et~al.(2019)Helber, Bischke, Dengel, and
  Borth}]{helber2019eurosat}
Patrick Helber, Benjamin Bischke, Andreas Dengel, and Damian Borth. 2019.
\newblock Eurosat: A novel dataset and deep learning benchmark for land use and
  land cover classification.
\newblock \emph{IEEE Journal of Selected Topics in Applied Earth Observations
  and Remote Sensing}, 12(7):2217--2226.

\bibitem[{Hendrycks et~al.(2021{\natexlab{a}})Hendrycks, Basart, Mu, Kadavath,
  Wang, Dorundo, Desai, Zhu, Parajuli, Guo et~al.}]{hendrycks2021many}
Dan Hendrycks, Steven Basart, Norman Mu, Saurav Kadavath, Frank Wang, Evan
  Dorundo, Rahul Desai, Tyler Zhu, Samyak Parajuli, Mike Guo, et~al.
  2021{\natexlab{a}}.
\newblock The many faces of robustness: A critical analysis of
  out-of-distribution generalization.
\newblock In \emph{Proceedings of the IEEE/CVF International Conference on
  Computer Vision}, pages 8340--8349.

\bibitem[{Hendrycks et~al.(2021{\natexlab{b}})Hendrycks, Zhao, Basart,
  Steinhardt, and Song}]{hendrycks2021natural}
Dan Hendrycks, Kevin Zhao, Steven Basart, Jacob Steinhardt, and Dawn Song.
  2021{\natexlab{b}}.
\newblock Natural adversarial examples.
\newblock In \emph{Proceedings of the IEEE/CVF Conference on Computer Vision
  and Pattern Recognition}, pages 15262--15271.

\bibitem[{Hinton et~al.(2015)Hinton, Vinyals, and Dean}]{hinton2015distilling}
Geoffrey Hinton, Oriol Vinyals, and Jeff Dean. 2015.
\newblock Distilling the knowledge in a neural network.
\newblock \emph{stat}, 1050:9.

\bibitem[{Huo et~al.(2021)Huo, Zhang, Liu, Lu, Gao, Yang, Wen, Zhang, Xu, Zheng
  et~al.}]{huo2021wenlan}
Yuqi Huo, Manli Zhang, Guangzhen Liu, Haoyu Lu, Yizhao Gao, Guoxing Yang,
  Jingyuan Wen, Heng Zhang, Baogui Xu, Weihao Zheng, et~al. 2021.
\newblock Wenlan: Bridging vision and language by large-scale multi-modal
  pre-training.
\newblock \emph{arXiv preprint arXiv:2103.06561}.

\bibitem[{Ilharco et~al.(2021)Ilharco, Wortsman, Wightman, Gordon, Carlini,
  Taori, Dave, Shankar, Namkoong, Miller, Hajishirzi, Farhadi, and
  Schmidt}]{ilharco_gabriel_2021_5143773}
Gabriel Ilharco, Mitchell Wortsman, Ross Wightman, Cade Gordon, Nicholas
  Carlini, Rohan Taori, Achal Dave, Vaishaal Shankar, Hongseok Namkoong, John
  Miller, Hannaneh Hajishirzi, Ali Farhadi, and Ludwig Schmidt. 2021.
\newblock \href {https://doi.org/10.5281/zenodo.5143773} {Openclip}.
\newblock If you use this software, please cite it as below.

\bibitem[{Kay et~al.(2017)Kay, Carreira, Simonyan, Zhang, Hillier,
  Vijayanarasimhan, Viola, Green, Back, Natsev et~al.}]{kay2017kinetics}
Will Kay, Joao Carreira, Karen Simonyan, Brian Zhang, Chloe Hillier, Sudheendra
  Vijayanarasimhan, Fabio Viola, Tim Green, Trevor Back, Paul Natsev, et~al.
  2017.
\newblock The kinetics human action video dataset.
\newblock \emph{arXiv preprint arXiv:1705.06950}.

\bibitem[{Ko and Gu(2022)}]{ko2022large}
Byungsoo Ko and Geonmo Gu. 2022.
\newblock Large-scale bilingual language-image contrastive learning.
\newblock \emph{arXiv preprint arXiv:2203.14463}.

\bibitem[{Krause et~al.(2013)Krause, Stark, Deng, and Fei-Fei}]{Krause2013}
Jonathan Krause, Michael Stark, Jia Deng, and Li~Fei-Fei. 2013.
\newblock 3d object representations for fine-grained categorization.
\newblock In \emph{4th International IEEE Workshop on 3D Representation and
  Recognition (3dRR-13)}, Sydney, Australia.

\bibitem[{Krizhevsky et~al.(2009)Krizhevsky, Hinton
  et~al.}]{krizhevsky2009learning}
Alex Krizhevsky, Geoffrey Hinton, et~al. 2009.
\newblock Learning multiple layers of features from tiny images.

\bibitem[{Lan et~al.(2017)Lan, Li, and Dong}]{lan2017fluency}
Weiyu Lan, Xirong Li, and Jianfeng Dong. 2017.
\newblock Fluency-guided cross-lingual image captioning.
\newblock In \emph{Proceedings of the 25th ACM international conference on
  Multimedia}, pages 1549--1557.

\bibitem[{Li et~al.(2019)Li, Xu, Wang, Lan, Jia, Yang, and Xu}]{li2019coco}
Xirong Li, Chaoxi Xu, Xiaoxu Wang, Weiyu Lan, Zhengxiong Jia, Gang Yang, and
  Jieping Xu. 2019.
\newblock Coco-cn for cross-lingual image tagging, captioning, and retrieval.
\newblock \emph{IEEE Transactions on Multimedia}, 21(9):2347--2360.

\bibitem[{Lin et~al.(2014)Lin, Maire, Belongie, Hays, Perona, Ramanan,
  Doll{\'a}r, and Zitnick}]{lin2014microsoft}
Tsung-Yi Lin, Michael Maire, Serge Belongie, James Hays, Pietro Perona, Deva
  Ramanan, Piotr Doll{\'a}r, and C~Lawrence Zitnick. 2014.
\newblock Microsoft coco: Common objects in context.
\newblock In \emph{European conference on computer vision}, pages 740--755.
  Springer.

\bibitem[{Nilsback and Zisserman(2008)}]{nilsback2008automated}
Maria-Elena Nilsback and Andrew Zisserman. 2008.
\newblock Automated flower classification over a large number of classes.
\newblock In \emph{2008 Sixth Indian Conference on Computer Vision, Graphics \&
  Image Processing}, pages 722--729. IEEE.

\bibitem[{Parkhi et~al.(2012)Parkhi, Vedaldi, Zisserman, and
  Jawahar}]{parkhi2012cats}
Omkar~M Parkhi, Andrea Vedaldi, Andrew Zisserman, and CV~Jawahar. 2012.
\newblock Cats and dogs.
\newblock In \emph{2012 IEEE conference on computer vision and pattern
  recognition}, pages 3498--3505. IEEE.

\bibitem[{Portaz et~al.(2019)Portaz, Randrianarivo, Nivaggioli, Maudet, Servan,
  and Peyronnet}]{portaz2019image}
Maxime Portaz, Hicham Randrianarivo, Adrien Nivaggioli, Estelle Maudet,
  Christophe Servan, and Sylvain Peyronnet. 2019.
\newblock Image search using multilingual texts: a cross-modal learning
  approach between image and text.
\newblock \emph{arXiv preprint arXiv:1903.11299}.

\bibitem[{Radford et~al.(2021)Radford, Kim, Hallacy, Ramesh, Goh, Agarwal,
  Sastry, Askell, Mishkin, Clark et~al.}]{radford2021learning}
Alec Radford, Jong~Wook Kim, Chris Hallacy, Aditya Ramesh, Gabriel Goh,
  Sandhini Agarwal, Girish Sastry, Amanda Askell, Pamela Mishkin, Jack Clark,
  et~al. 2021.
\newblock Learning transferable visual models from natural language
  supervision.
\newblock In \emph{International Conference on Machine Learning}, pages
  8748--8763. PMLR.

\bibitem[{Recht et~al.(2019)Recht, Roelofs, Schmidt, and
  Shankar}]{recht2019imagenet}
Benjamin Recht, Rebecca Roelofs, Ludwig Schmidt, and Vaishaal Shankar. 2019.
\newblock Do imagenet classifiers generalize to imagenet?
\newblock In \emph{International Conference on Machine Learning}, pages
  5389--5400. PMLR.

\bibitem[{Rombach et~al.(2022)Rombach, Blattmann, Lorenz, Esser, and
  Ommer}]{rombach2022high}
Robin Rombach, Andreas Blattmann, Dominik Lorenz, Patrick Esser, and Bj{\"o}rn
  Ommer. 2022.
\newblock High-resolution image synthesis with latent diffusion models.
\newblock In \emph{Proceedings of the IEEE/CVF Conference on Computer Vision
  and Pattern Recognition}, pages 10684--10695.

\bibitem[{Schuhmann et~al.(2022)Schuhmann, Beaumont, Vencu, Gordon, Wightman,
  Cherti, Coombes, Katta, Mullis, Wortsman et~al.}]{schuhmann2022laion}
Christoph Schuhmann, Romain Beaumont, Richard Vencu, Cade Gordon, Ross
  Wightman, Mehdi Cherti, Theo Coombes, Aarush Katta, Clayton Mullis, Mitchell
  Wortsman, et~al. 2022.
\newblock Laion-5b: An open large-scale dataset for training next generation
  image-text models.
\newblock \emph{arXiv preprint arXiv:2210.08402}.

\bibitem[{Schuhmann et~al.(2021)Schuhmann, Vencu, Beaumont, Kaczmarczyk,
  Mullis, Katta, Coombes, Jitsev, and Komatsuzaki}]{schuhmann2021laion}
Christoph Schuhmann, Richard Vencu, Romain Beaumont, Robert Kaczmarczyk,
  Clayton Mullis, Aarush Katta, Theo Coombes, Jenia Jitsev, and Aran
  Komatsuzaki. 2021.
\newblock Laion-400m: Open dataset of clip-filtered 400 million image-text
  pairs.
\newblock \emph{arXiv preprint arXiv:2111.02114}.

\bibitem[{Sharma et~al.(2018)Sharma, Ding, Goodman, and
  Soricut}]{sharma2018conceptual}
Piyush Sharma, Nan Ding, Sebastian Goodman, and Radu Soricut. 2018.
\newblock Conceptual captions: A cleaned, hypernymed, image alt-text dataset
  for automatic image captioning.
\newblock In \emph{Proceedings of the 56th Annual Meeting of the Association
  for Computational Linguistics (Volume 1: Long Papers)}, pages 2556--2565.

\bibitem[{Soomro et~al.(2012)Soomro, Zamir, and Shah}]{soomro2012ucf101}
Khurram Soomro, Amir~Roshan Zamir, and Mubarak Shah. 2012.
\newblock Ucf101: A dataset of 101 human actions classes from videos in the
  wild.
\newblock \emph{arXiv preprint arXiv:1212.0402}.

\bibitem[{Stallkamp et~al.(2011)Stallkamp, Schlipsing, Salmen, and
  Igel}]{stallkamp2011german}
Johannes Stallkamp, Marc Schlipsing, Jan Salmen, and Christian Igel. 2011.
\newblock The german traffic sign recognition benchmark: a multi-class
  classification competition.
\newblock In \emph{The 2011 international joint conference on neural networks},
  pages 1453--1460. IEEE.

\bibitem[{Tiedemann(2012)}]{tiedemann2012parallel}
J{\"o}rg Tiedemann. 2012.
\newblock Parallel data, tools and interfaces in opus.
\newblock In \emph{Lrec}, volume 2012, pages 2214--2218.

\bibitem[{Veeling et~al.(2018)Veeling, Linmans, Winkens, Cohen, and
  Welling}]{veeling2018rotation}
Bastiaan~S Veeling, Jasper Linmans, Jim Winkens, Taco Cohen, and Max Welling.
  2018.
\newblock Rotation equivariant cnns for digital pathology.
\newblock In \emph{International Conference on Medical image computing and
  computer-assisted intervention}, pages 210--218. Springer.

\bibitem[{Wang et~al.(2019)Wang, Ge, Lipton, and Xing}]{wang2019learning}
Haohan Wang, Songwei Ge, Zachary Lipton, and Eric~P Xing. 2019.
\newblock Learning robust global representations by penalizing local predictive
  power.
\newblock \emph{Advances in Neural Information Processing Systems}, 32.

\bibitem[{Wang et~al.(2022)Wang, Zhang, Zhang, Yang, Gao, Wu, Dong, He, Zhuo,
  Yang, Huang, Li, Wu, Lu, Zhu, Chen, Han, Pan, Wang, Wang, Wu, Zeng, Chen,
  Gan, and Zhang}]{fengshenbang}
Junjie Wang, Yuxiang Zhang, Lin Zhang, Ping Yang, Xinyu Gao, Ziwei Wu, Xiaoqun
  Dong, Junqing He, Jianheng Zhuo, Qi~Yang, Yongfeng Huang, Xiayu Li, Yanghan
  Wu, Junyu Lu, Xinyu Zhu, Weifeng Chen, Ting Han, Kunhao Pan, Rui Wang, Hao
  Wang, Xiaojun Wu, Zhongshen Zeng, Chongpei Chen, Ruyi Gan, and Jiaxing Zhang.
  2022.
\newblock Fengshenbang 1.0: Being the foundation of chinese cognitive
  intelligence.
\newblock \emph{CoRR}, abs/2209.02970.

\bibitem[{{Xiao} et~al.(2010){Xiao}, {Hays}, {Ehinger}, {Oliva}, and
  {Torralba}}]{Xiao2010}
J.~{Xiao}, J.~{Hays}, K.~A. {Ehinger}, A.~{Oliva}, and A.~{Torralba}. 2010.
\newblock \href {https://doi.org/10.1109/CVPR.2010.5539970} {Sun database:
  Large-scale scene recognition from abbey to zoo}.
\newblock In \emph{2010 IEEE Computer Society Conference on Computer Vision and
  Pattern Recognition}, pages 3485--3492.

\bibitem[{Xie et~al.(2022)Xie, Cai, Song, Li, Kong, Wu, Morimitsu, Yao, Wang,
  Leng et~al.}]{xie2022zero}
Chunyu Xie, Heng Cai, Jianfei Song, Jincheng Li, Fanjing Kong, Xiaoyu Wu,
  Henrique Morimitsu, Lin Yao, Dexin Wang, Dawei Leng, et~al. 2022.
\newblock Zero and r2d2: A large-scale chinese cross-modal benchmark and a
  vision-language framework.
\newblock \emph{arXiv preprint arXiv:2205.03860}.

\bibitem[{Xu(2019)}]{xu2019nlp}
Bright Xu. 2019.
\newblock Nlp chinese corpus: Large scale chinese corpus for nlp.

\bibitem[{Yang et~al.(2022)Yang, Pan, Lin, Men, Zhang, Zhou, and
  Zhou}]{yang2022chinese}
An~Yang, Junshu Pan, Junyang Lin, Rui Men, Yichang Zhang, Jingren Zhou, and
  Chang Zhou. 2022.
\newblock Chinese clip: Contrastive vision-language pretraining in chinese.
\newblock \emph{arXiv preprint arXiv:2211.01335}.

\bibitem[{Yang et~al.(2020)Yang, Cer, Ahmad, Guo, Law, Constant,
  Hernandez~Abrego, Yuan, Tar, Sung, Strope, and
  Kurzweil}]{yang-etal-2020-multilingual}
Yinfei Yang, Daniel Cer, Amin Ahmad, Mandy Guo, Jax Law, Noah Constant, Gustavo
  Hernandez~Abrego, Steve Yuan, Chris Tar, Yun-hsuan Sung, Brian Strope, and
  Ray Kurzweil. 2020.
\newblock \href {https://doi.org/10.18653/v1/2020.acl-demos.12} {Multilingual
  universal sentence encoder for semantic retrieval}.
\newblock In \emph{Proceedings of the 58th Annual Meeting of the Association
  for Computational Linguistics: System Demonstrations}, pages 87--94, Online.
  Association for Computational Linguistics.

\bibitem[{Young et~al.(2014)Young, Lai, Hodosh, and
  Hockenmaier}]{young2014image}
Peter Young, Alice Lai, Micah Hodosh, and Julia Hockenmaier. 2014.
\newblock From image descriptions to visual denotations: New similarity metrics
  for semantic inference over event descriptions.
\newblock \emph{Transactions of the Association for Computational Linguistics},
  2:67--78.

\bibitem[{Yuan et~al.(2021)Yuan, Zhao, Du, Ding, Liu, Cen, Zou, Yang, and
  Tang}]{yuan2021wudaocorpora}
Sha Yuan, Hanyu Zhao, Zhengxiao Du, Ming Ding, Xiao Liu, Yukuo Cen, Xu~Zou,
  Zhilin Yang, and Jie Tang. 2021.
\newblock Wudaocorpora: A super large-scale chinese corpora for pre-training
  language models.
\newblock \emph{AI Open}, 2:65--68.

\bibitem[{Zhai et~al.(2022)Zhai, Wang, Mustafa, Steiner, Keysers, Kolesnikov,
  and Beyer}]{zhai2022lit}
Xiaohua Zhai, Xiao Wang, Basil Mustafa, Andreas Steiner, Daniel Keysers,
  Alexander Kolesnikov, and Lucas Beyer. 2022.
\newblock Lit: Zero-shot transfer with locked-image text tuning.
\newblock In \emph{Proceedings of the IEEE/CVF Conference on Computer Vision
  and Pattern Recognition}, pages 18123--18133.

\end{thebibliography}
